\pdfoutput=1
\documentclass[11pt]{article}

\usepackage[final]{acl}

\usepackage{times}
\usepackage{latexsym}
\usepackage{graphicx}
\usepackage[T1]{fontenc}
\usepackage[utf8]{inputenc}

\usepackage{microtype}

\usepackage{inconsolata}
\usepackage{bbm}
\usepackage{amsmath}
\usepackage{booktabs}
\usepackage{algorithm}
\usepackage{algpseudocode}

\algrenewcommand{\algorithmiccomment}[1]{\hfill// #1}
\usepackage{multirow}

\title{Arithmetic Reasoning with LLM: Prolog Generation \& Permutation}
\author{Xiaocheng Yang \and Bingsen Chen \and Yik-Cheung Tam\\
  Shanghai Frontiers Science Center of Artificial Intelligence and Deep Learning\\
  New York University Shanghai\\
  \{\href{mailto:xy2128@nyu.edu}{xy2128},\href{mailto:bc3088@nyu.edu}{bc3088},\href{mailto:yt2267@nyu.edu}{yt2267}\}@nyu.edu}
\begin{document}
\maketitle
\begin{abstract}
Instructing large language models (LLMs) to solve elementary school math problems has shown great success using Chain of Thought (CoT). However, the CoT approach relies on an LLM to generate a sequence of arithmetic calculations which can be prone to cascaded calculation errors. We hypothesize that an LLM should focus on extracting predicates and generating symbolic formulas from the math problem description so that the underlying calculation can be done via an external code interpreter.
We investigate using LLM to generate Prolog programs to solve mathematical questions. Experimental results show that our Prolog-based arithmetic problem-solving outperforms CoT generation in the GSM8K benchmark across three distinct LLMs. In addition, given the insensitive ordering of predicates and symbolic formulas in Prolog, we propose to permute the ground truth predicates for more robust LLM training via data augmentation.

\end{abstract}

\section{Introduction}

Large language models (LLMs), with their scaling of model size and data size, have demonstrated impressive performance across various understanding and generation tasks \citep{gpt3, chowdhery2022palm, gopher, lambda, touvron2023llama, falcon40b, jiang2023mistral}. Nevertheless, such LLMs fall short in addressing mathematical problems that involves arithmetic, commonsense, and symbolic reasoning – topics that may appear deceptively simple to humans \citep{gopher}. Existing works leveraged Chain-of-Thought (CoT) reasoning that asks language models to generate both the answer and the step-by-step reasoning chain, which helps break down a complex reasoning task into a sequential thought process \cite{wei2022chain}. Particularly, arithmetic reasoning with CoT is shown to be an emergent ability that language models acquired during the scaling process \citep{wei2022emergent}. 

\begin{figure}[H]
    \centering
    \includegraphics[width=\linewidth]{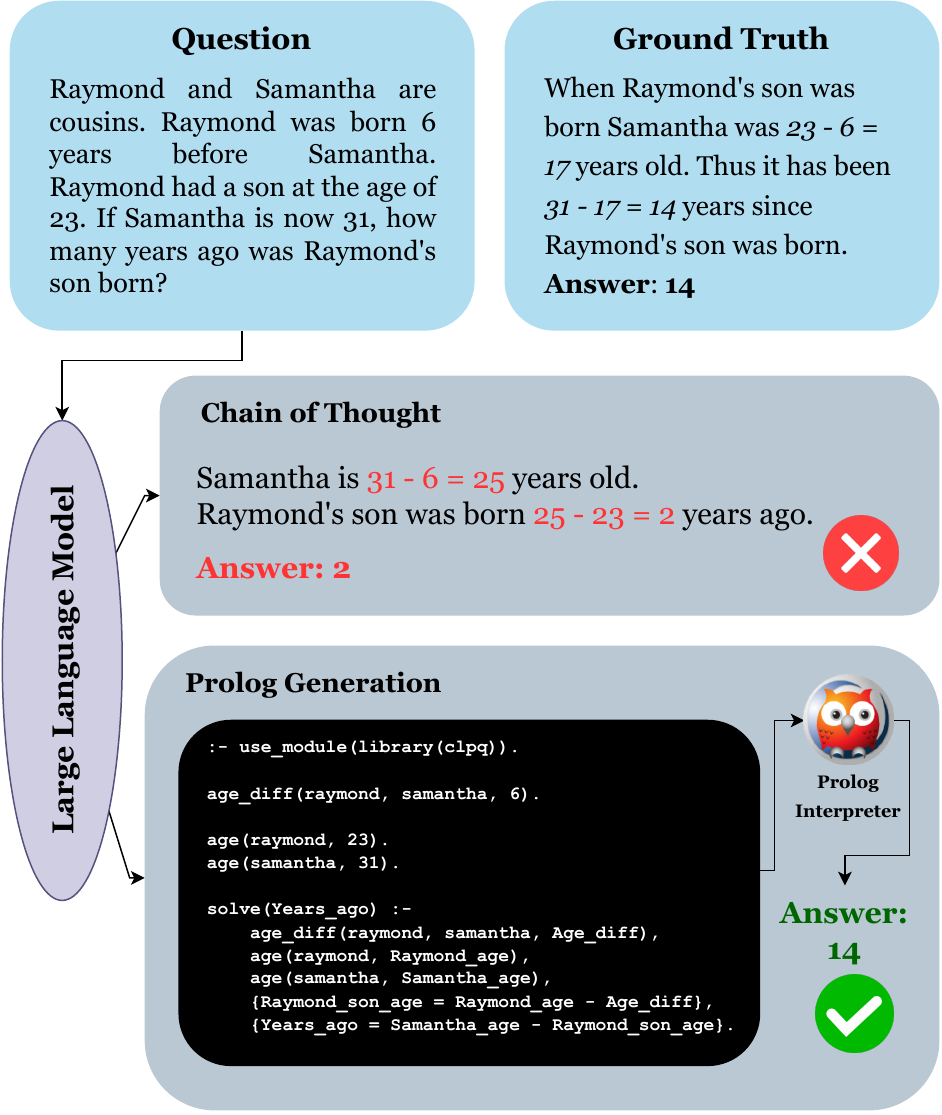}
    \caption{Overview of Prolog generation for arithmetic reasoning with large language models.}
    \label{fig:enter-label}
\end{figure}

Yet, natural language reasoning is not native to mathematical operations and symbolic manipulations. A line of work has focused on augmenting language models with deterministic computation resources like a calculator \citep{schick2023toolformer} or program-based tools \citep{gao2023pal, gou2023tora}. However, all such methods require a sequential reasoning trajectory, where models need to translate the natural language questions into sequential mathematical or logical operations. Our research probes into the application of Prolog, a logic programming language, in solving the arithmetic reasoning task. Prolog solves arithmetic reasoning tasks by defining an unordered set of predicates and running queries over them. We further explain the unique properties of Prolog in Section~\ref{sec:prolog}. In Prolog code generation for arithmetic reasoning, LLMs extract facts and rules in mathematical questions and formulate them into Prolog code. If the facts and rules are accurately captured, a Prolog interpreter can precisely solve for a correct answer in a deterministic way.

Our research has the following contributions: 1) We curate and open-source the GSM8K-Prolog dataset with a semi-automatic approach, which contains arithmetic reasoning problems and their corresponding Prolog code solutions. 2) Our experiments show that Prolog code generation is consistently better than CoT on the arithmetic reasoning task, indicating that LLM can focus on predicate extractions and rely on an external tool to calculate and perform the logical induction to address mathematical problems. 3) Given the non-sequential nature of predicates in Prolog code, we propose predicate permutation as a data augmentation method and demonstrate its efficacy in robust LLM training.

\begin{figure}[H]
    \centering
    \includegraphics[width=\linewidth]{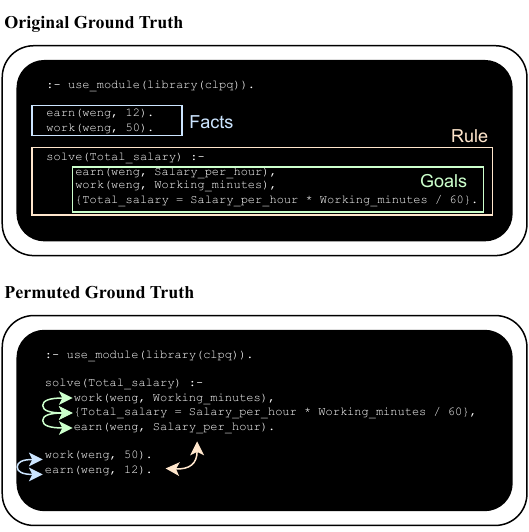}
    \caption{Prolog and permuted Prolog code samples.}
    \label{fig:prolog_sample}
\end{figure}

\section{Preliminaries: Prolog Language} \label{sec:prolog}
Prolog is a logic programming language, which was initially designed for artificial intelligence and computational linguistics \cite{Clocksin_Mellish_2003,Bratko_2012,Covington_2002}. As shown in the upper graph of Figure~\ref{fig:prolog_sample}, a Prolog program defines a set of predicates that contains facts and goals. In the example, facts include \texttt{earn(weng, 12)} that declares the hourly salary of Weng, and \texttt{work(weng, 50)} that defines the working minutes of Weng; the goals constitute a rule in the form of \texttt{solve<answer>:-<goal\_1>,<goal\_2>, ...}. A rule is true when all the goals are satisfied. Having all the facts and goals defined in the program, users can make a query to obtain the solutions that make the rule true given all the facts. Moreover, Prolog codes are not sequential like Python, meaning that the order of facts and rules does not alter the result of the program. The lower graph in Figure~\ref{fig:prolog_sample}, shows an equivalent sample that permutes the order of the predicates, which produces the same result as the original program.

\addtolength{\tabcolsep}{0.2em}
\begin{table*}[t]
    \centering
    \small
    \renewcommand{\arraystretch}{1.25}
    \begin{tabular}{l|c c c c c c}
    \toprule
        \multirow{2}{*}{Method} & \multicolumn{2}{c}{\textbf{Llama-2}} & \multicolumn{2}{c}{\textbf{CodeLlama}} & \multicolumn{2}{c}{\textbf{Mistral}} \\ 
        & GSM8K & GSM-HARD & GSM8K & GSM-HARD & GSM8K & GSM-HARD \\
        \midrule
        CoT & 33.8\% & 12.0\% & 37.5\% & 13.9\% & 58.9\% & 30.8\%\\
        \midrule
        Prolog & 41.5\% & 32.4\% & 55.0\% & 41.6\% & 66.3\% & 50.6\% \\
        \textsc{ProPer} & \textbf{51.0\%} & \textbf{37.4\%} & \textbf{59.0\%} & \textbf{45.9\%} & \textbf{70.2\%} & \textbf{54.4\%} \\
    \bottomrule
    \end{tabular} 
    \caption{Accuracy results on the GSM8K and GSM-HARD datasets. We compare regular Prolog generation (Prolog) and \textsc{ProPer} Prolog generation with the CoT baseline (supervised finetuning with LoRA using CoT ground truth labels in the original GSM8K dataset). }
    \label{tab:main}
\end{table*}

\section{Method}
\subsection{GSM8K-Prolog Dataset} \label{sec: GSM8K-Prolog Dataset}
To our knowledge, there has not been a dataset for solving mathematical questions with Prolog. We hence curated a dataset based on GSM8K \citep{cobbe2021training}, a popular benchmark of diverse grade school math word problems, in a semi-automatic manner with OpenAI's Text Completion API \footnote{\url{https://platform.openai.com/docs/guides/text-generation/chat-completions-api}}. In particular, we used the same dataset splits and questions in GSM8K and prompted GPT-4 to generate the Prolog programs to solve the questions. We then manually corrected some malfunctioning samples. In this manner, we obtained a high-quality corpus with $100\%$ accuracy in terms of the code results. Algorithm~\ref{alg:cap} describes the detailed pseudo-code for creating this dataset. We open-sourced this dataset to the research community with the MIT license. \footnote{\url{https://huggingface.co/datasets/Thomas-X-Yang/gsm8k-prolog}}.
% Contribution 1

% Main results table
\subsection{\textsc{ProPer}: \underline{Pro}log \underline{Per}mutation} \label{sec:permutation}
Since Prolog predicates are permutable, inspired by XLNet \cite{yang2020xlnet} that performs a token-wise permutation via attention masking, we decided to also use the permutation technique. The XLNet, via the permutation, can attend to tokens on both sides during training and thus can partially obtain the property of autoencoding while maintaining the property of autoregressive modeling. Similarly, \textsc{ProPer} takes advantage of the permutative property of facts and goals in the Prolog programs as indicated in Figure~\ref{fig:prolog_sample}. For each original program, we sample $n$ of its permutations and mix them into the dataset. In this way, models can learn to extract predicates in the mathematical questions based on any other predicates regardless of the ordering, which more precisely reflects the nature of the Prolog language. We describe the practical details of permutation in Appendix~\ref{apdx:permute}.

\section{Experiments}

\subsection{Setup}
% Instruction template, training configuration, evaluation metrics ...
\paragraph{Dataset}
We used the GSM8K-Prolog described in Section~\ref{sec: GSM8K-Prolog Dataset}. We denote the corpus as $D$. The training set is $D_{train}$ and the test set is $D_{test}$. The total corpus size is 8792, where 7473 samples belong to the training set and 1319 belong to the test set. During training, 100 samples were selected from the training set to constitute the validation set. The input format follows the instruction prompt used in Stanford Alpaca \cite{alpaca} (See sample prompts in Appendix~\ref{sec:ins_prompt}). We discarded samples that exceeded 512 tokens. Notably, when we used \textsc{ProPer} to augment the dataset, we used slightly altered input prompts for permuted samples because we found that using the same instruction for both the original ground truth codes and the permuted ones degraded the performance of the model. A likely reason is that having multiple correct output tokens for the same input instruction confuses the model. In addition, besides the GSM8K's test set, GSM-HARD \cite{gao2023pal}, which replaces the numbers in the GSM8K test set with large numbers and thus makes questions hard for language models, was also used for evaluation.

\paragraph{Training}
We experimented with different LLMs' 7B versions, including Llama2 \cite{touvron2023llama}, CodeLlama \cite{rozière2023code} and Mistral \cite{jiang2023mistral}. We adopted 8-bit quantization and LoRA \cite{hu2021lora} to finetune models efficiently at a reasonable performance cost. We applied LoRA to finetune query and value weight matrices in the transformer blocks. We experimented with different LoRA rank and alpha settings, including $(r, \alpha)= (8, 16), (16, 32)$, and $(32, 64)$. With more trainable parameters, $r=32, \alpha=64$ yielded significantly better results, which we thereby adopted as the configuration for all the experiments. Note that this setting resulted in training only 0.248\% of the 7 billion parameters for Llama2 and CodeLlama, and 0.188\% of the 7 billion for Mistral. We document our training details and GPU usage in Appendix~\ref{sec:gpu}.

\paragraph{Evaluation}
At inference time, we used beam search with a beam size of 4 to generate the Prolog code. We then used the PySwip library \footnote{Prolog version 9.0.4. PySwip version 0.2.11. \url{https://github.com/yuce/pyswip}}, a foreign interface of Prolog in Python, as the Prolog interpreter to produce the final answer. We used accuracy as the metric for evaluation. It is defined as 
\[\text{Acc} = \frac{\sum_{i=1}^{|D_{test}|} \mathbbm{1}_{\left\{\mathcal{P}(a^{pred}_i)=\mathcal{P}(a^{true}_i)\right\}}}{|D_{test}|}\times 100\%\]
where $\mathcal{P}$ denotes the Prolog interpreter. Notably, since we noticed that the PySwip library cannot handle decimal answers, we only considered the samples with an integer answer.

% Besides, we keep a record of syntax error rates and semantic error rates. The syntax error rate is defined as
% \[\text{SynErr} = \frac{\sum_{i=1}^{|D_{test}|} \mathbbm{1}_{\left\{\mathcal{P}(a^{pred}_i)\text{ raises exception}\right\}}}{|D_{test}|}\]
% and the semantics error rate is defined as

% \[\text{SemErr} = \frac{\sum_{i=1}^{|D_{test}|} \mathbbm{1}_{\left\{\mathcal{P}(a^{pred}_i)\neq\mathcal{P}(a^{true}_i)\right\}}}{|D_{test}|}\]

\subsection{Results}

\begin{figure*}[t]
    \begin{minipage}{0.4\textwidth}
        \centering
        \footnotesize
        \renewcommand{\arraystretch}{1.5}
        \addtolength{\tabcolsep}{-0.5em}
        \begin{tabular}{c|ccc}
            \toprule
            Ratio &  \textbf{Llama-2}  & \textbf{CodeLlama}  &  \textbf{Mistral}  \\
            \midrule
            1:0          &    41.5&  55.0&  66.3\\
            1:1          &    50.9 (49.5)&  58.7 (56.6)&  \textbf{70.2 (69.1)}\\
            1:2          &    \textbf{51.0 (49.4)}&  \textbf{59.0 (58.3)}&  68.8 (66.8) \\
            \bottomrule
        \end{tabular}
        % \addtolength{\tabcolsep}{0.2em}
        \captionof{table}{Accuracy(\%) results on GSM8K with different permutation ratios. We report both the best and average accuracy of \texttt{1:1} and \texttt{1:2} over three trials with different randomly permuted data in the form of \texttt{max} (\texttt{avg}). Note that the \texttt{1:0} case essentially means not applying \textsc{ProPer}.}
        \label{table:proper acc}
    \end{minipage}%
    \hspace{3mm}
    \begin{minipage}{0.6\textwidth}
        \centering
        \includegraphics[width=\linewidth]{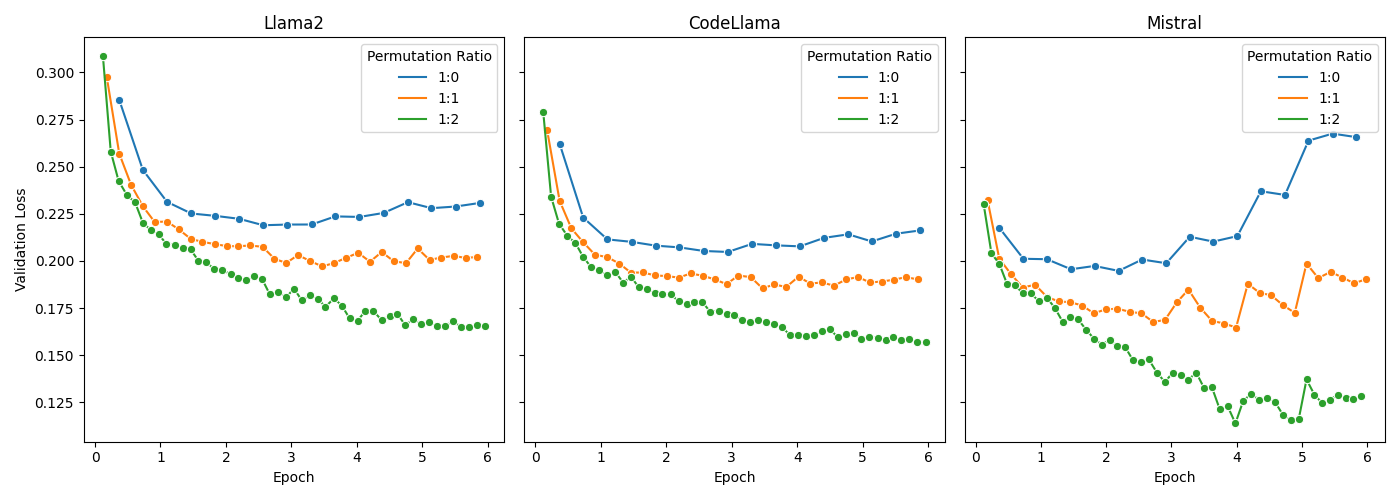}
        \captionof{figure}[width=.9\linewidth]{Validation loss curves for training Llama2, CodeLlama, and Mistral with different permutation ratios (We only report the first trial when we use permuted data since the loss curves are very similar across trials).}
        \label{fig:proper loss}
    \end{minipage}
\end{figure*}

\paragraph{Prolog generation performs consistently better than CoT across three models.}
According to Table~\ref{tab:main}, generating Prolog to solve mathematical questions yields significantly more accurate results with a 10.9\% margin over the CoT baseline on average across all models on GSM8K. This gap further expands to 22.6\% on GSM-HARD, indicating exceptional superiority over CoT when large number calculations are involved. Although Llama-2 and Mistral exhibit large performance gaps when applying CoT reasoning, generating Prolog code produces better results than CoT on both models. This observation indicates that Prolog generation works well regardless of the model's inherent arithmetic reasoning capability. Also, CodeLlama demonstrates a larger performance gain when switching from CoT to Prolog generation, which is potentially attributed to its pretraining on the code-related corpus. In other words, CodeLlama is specifically trained to generate structured programs better than natural language reasoning.

\paragraph{With a proper permutation ratio, \textsc{ProPer} further enhances LLM's arithmetic reasoning with Prolog generation.} Permutation ratio refers to the ratio between original samples and permuted samples. As shown in Table~\ref{table:proper acc}, by adding two permuted samples for each original sample, we observed an increased accuracy of 9.5\% and 4.0\% of Llama-2 and CodeLlama respectively on the test set. This improvement indicates that learning the non-sequential structure of Prolog predicates is helpful for LLMs to generate correct Prolog programs to solve arithmetic problems. On the other hand, the lowered accuracy of Mistral, compared with its case of one permutation per sample, suggests that \textsc{ProPer} might be limited for models already with high Prolog generation capacity.

\paragraph{Lowered validation loss from \textsc{ProPer} does not lead to higher accuracy.}
As is shown in Figure~\ref{fig:proper loss}, increasing the permutation ratio results in significantly lowered validation loss. This is because we first added in permutations and then split a validation set from the training set. Consequently, the permutations of validation samples were included in the training set and the generalization ability of the language models enabled the models to utilize the permutations to improve the performance on the validation set, causing a soft data leakage. Therefore, according to Table~\ref{table:proper acc}, the permutation ratio of \texttt{1:2} yielded a weakened performance on Mistral although the validation loss was the lowest.

\begin{figure}[h]
    \centering
    \includegraphics[width=\linewidth]{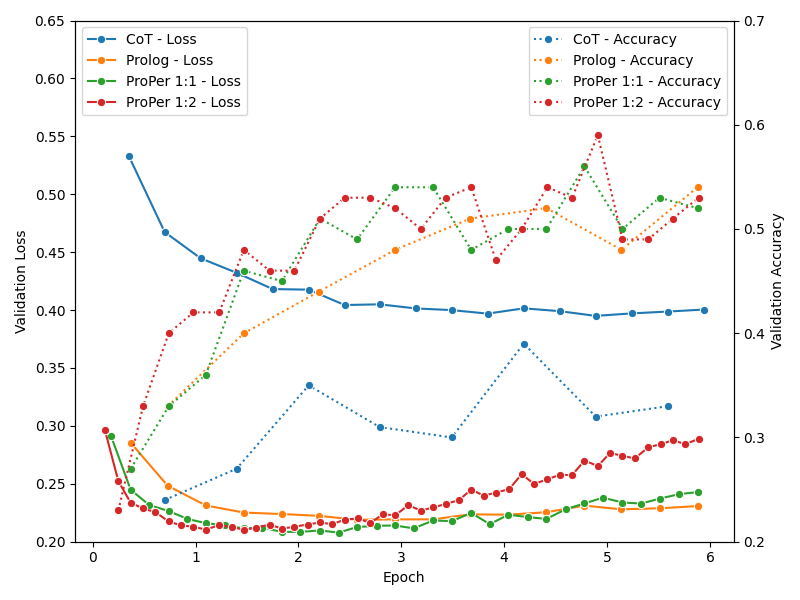}
    \caption{Validation loss curves and validation accuracy curves for training Llama2 with different methods (We only report the first trial when we use permuted data since the loss and accuracy curves are very similar across trials).}
    \label{fig:acc check}
\end{figure}

\addtolength{\tabcolsep}{0.2em}
\begin{table}[h]
    \centering
    \small
    \renewcommand{\arraystretch}{1.25}
    \addtolength{\tabcolsep}{-0.5em}
    \begin{tabular}{c|ccc}
            \toprule
            \multirow{2}{*}{Method} &  \multirow{2}{*}{\textbf{Initial}}  & \textbf{No Leakage}  &  \textbf{No Leakage} \\
            & & (by loss) & (by accuracy) \\
            \midrule
            CoT         &    33.8 & 33.8 & 36.5  \\
            Prolog      &    41.5 & 41.5 & 47.9  \\
            ProPer 1:1  &    50.9 (49.5) & 44.3 (43.4) & 50.1 (48.4)  \\
            ProPer 1:2  &    \textbf{51.0 (49.4)} & \textbf{44.4 (43.6)} & \textbf{51.3 (50.3)}  \\
            \bottomrule
        \end{tabular}
    \caption{Accuracy(\%) results of training Llama2 on the GSM8K dataset. We compare the results of avoiding validation sample leakage in the training set and picking the optimal checkpoint based on validation loss and accuracy with the initial results with leakage. The best and average accuracy of \texttt{1:1} and \texttt{1:2} are in the form of \texttt{max} (\texttt{avg}).}
    \label{tab:no leakage}
\end{table}

\paragraph{Increased validation loss from \textsc{ProPer} does not lead to decreased validation accuracy.}
Excluding the permutations of validation samples from the training set, we report both the cross-entropy loss and the accuracy on the validation set for Llama2 using different methods in Figure~\ref{fig:acc check}. A mismatch between the loss and accuracy is observed. As a loss curve decreases to the minimum and bounces back, the corresponding accuracy curve keeps increasing and then maintains a high level. As is shown in Table~\ref{tab:no leakage}, by choosing checkpoints based on validation accuracy instead of validation loss, the performance can be improved across all methods. Moreover, the improvement for the Prolog and \textsc{ProPer} method is significantly greater than that of CoT, suggesting a larger divergence between the objective of cross entropy loss and the ultimate accuracy of Prolog generation. Therefore, it is suggested to choose the best checkpoint based on the validation accuracy. Nevertheless, the new performance is similar to the initial results where leakage is involved. We notice that late checkpoints yield better performance according to the validation accuracy and the validation loss keeps decreasing in the initial setting. Therefore, both settings happen to pick late checkpoints, resulting in similar performance.

We have also tested Python generation, for which the corpus was generated by the same procedure as Algorithms~\ref{alg:cap} except that we prepare Python codes instead of Prolog codes. It gives an accuracy of 55.12\% on GSM8K using Llama2 as the base model, better than both Prolog and \textsc{ProPer}. One possible reason is that Python now is the prevalent programming language and Llama2 might have been pretrained on a large amount of Python codes. We believe if sufficient Prolog codes are used for training, Prolog generation can at least match up with Python generation due to its essence of symbolic reasoning.

We present some representative error cases of Mistral (\texttt{1:1}) in Appendix~\ref{sec:err}.

\section{Related Work}
\paragraph{Arithmetic Reasoning}
The Chain-of-Thought (CoT) prompting approach \cite{wei2022chain} first proposes to prompt the model to generate the reasoning chain step-by-step to reach the final answer. Afterwards, advancements have been made in LLMs' reasoning capacity via step-by-step methods \cite{zhou2023leasttomost, Zhu_2023, huang2022large, liang2023encouraging}. However, the natural language generation still performs poorly on complex or multi-step reasoning. Therefore, one trajectory of efforts has been made to leverage reasoning structures like trees \cite{yao2023tree, long2023large} and graphs \cite{besta2023graph, zhang2023cumulative}. Another trajectory is to render the reasoning task based on external tools \cite{cobbe2021training, mishra2023lila, gou2023tora, gao2023pal, pmlr-v202-shao23a, chen2023program}, which is the one that we are following. Besides, \citeposs{yuan2023scaling} RFT method shares the idea of dataset augmentation, but they compile rejection samples from multiple models to form an augmented training set, which is different from \textsc{ProPer}'s automatic permutation.

\paragraph{Neural Symbolic Reasoning}
Neural symbolic reasoning \cite{7780381, neelakantan2017learning, NEURIPS2019_c20a7ce2, Gupta2020Neural, nye2021improving} aims to leverage both neural networks and symbolic reasoning to obtain better reasoning abilities and transparency. Those methods suffer from low scalability of learning and reasoning components. LLMs are hence adopted to generate symbolic representations from natural language \cite{lyu2023faithful, pan2023logiclm, yang2023neurosymbolic}, where deterministic symbolic solvers will process the query and symbolic representations generated by LLMs to conduct reasoning or proofs. Prolog has been a popular candidate for the format of symbolic representations. We are posited on this trajectory and in the specific field of arithmetic reasoning.

\section{Conclusion}

In conclusion, we aim to enhance the reasoning performance of LLMs. We adopt the pipeline that the model generates Prolog predicates from a mathematical question in natural language and an external Prolog interpreter processes the query for a final result. We contribute an open-sourced corpus named GSM8K-Prolog, which is a high-quality Prolog-annotated version of GSM8K. We show that Prolog generation substantially outperformed CoT generation across all three 7B models for solving arithmetic reasoning problems. We also propose \textsc{ProPer}, a data augmentation method designed specifically for Prolog code generation, which enables the finetuned models to learn the non-sequential nature of Prolog predicates. \textsc{ProPer} further improves the model's accuracy on GSM8K-Prolog and mitigates early convergence during training. Lastly, due to the gap between cross-entropy loss objective and accuracy, we suggest using validation accuracy instead of validation loss to pick the best checkpoint.

\newpage
\section*{Limitations}
% Objective mismatch
% Float point counts as syntax error due to package limitation
Although we have experimentally conducted full-parameter finetuning, the result was not satisfying. We believe it is because of the limited size of the original corpus. Therefore, at the current stage, we cannot have a comparison with other methods like ToRA \citep{gou2023tora} or RFT \citep{yuan2023scaling}. Future research can look into preparing a larger and more diverse corpus adapted to Prolog code generation. Besides, We did not try scaling the base model to more than 7B parameters. So we do not know the impact of model scaling on the performance of Prolog code generation for arithmetic reasoning. Furthermore, due to the limitation of the PySwip library, solvable questions are restricted to the ones with an integer answer. Future work can expand the domain by using other interpreting tools.

\bibliography{anthology,custom}

\newpage
\onecolumn
\appendix
\section{Appendix}

\subsection{Generation Procedure of GSM8K-Prolog}
Below is the detailed pseudo-code for the GSM8K-Prolog dataset generation.

\begin{algorithm}
\caption{Procedure of GSM8K-Prolog Generation}\label{alg:cap}
\begin{algorithmic}
\Require The original GSM8K dataset, denoted as set $\mathcal{X} = \{(q_{i}, a^{\text{CoT}}_{i})\}_{i=1}^{N}$, where each sample consists of one question $q_{i}$ and one Chain-of-Thought answer $a^{\text{CoT}}_{i}$; A Prolog interpreter $\mathcal{P}$ that returns the output of a Prolog program; A Chain-of-Thought answer retriever $\mathcal{C}$ that parses out the final answer of a natural language reasoning chain.
\Ensure GSM8K-Prolog dataset $\mathcal{D} = \{(q_{i}, a^{\text{Prolog}}_{i})\}_{i=1}^{N}$
\\
\State Initialize a set of indices $\mathcal{I} \gets \{1,\cdots,N\}$, a static instruction prompt in the new dataset $p_{\text{ins}}$, and an initial question for querying OpenAI API $q_{\text{gen}}$.
\State Manually craft 10 correct Prolog codes $\{a_i^{\text{Prolog}}\}_{i=1}^{10}$ that correctly solve $\{q_i\}_{i=1}^{10}$ in $\mathcal{X}$ to initialize $\mathcal{D}$
\For{$i \in I$}
    \State Retrieve a sample $(q_{i}, a^{\text{CoT}}_{i}) \in \mathcal{X}$
    \State Prompt GPT-4 with $\{q^{gen}\} \cup \{(q_k, a_k^{\text{CoT}}, a_k^{\text{Prolog}})_{k=1}^{10}\} \cup \{ q_{i}, a^{\text{CoT}}_{i}\}$ to obtain $a_i^{\text{Prolog}}$
    \If{$\mathcal{P}(a_i^{\text{Prolog}}) = \mathcal{C}(a^{\text{CoT}}_{i})$}
        \State $\mathcal{D} \gets \mathcal{D} \cup \{(p_{\text{ins}}, q_{i}, a_i^{\text{Prolog}})\}$
        \State $\mathcal{I} \gets \mathcal{I} \setminus \{i\}$
    \EndIf
\EndFor

\State Manually select the top 10 clean and logical Prolog code from the current $\mathcal{D}$ to form a new few-shot sample set $Q^{\text{fixed}} = \{(q_k, a_k^{\text{CoT}}, a_k^{\text{Prolog}})_{k\notin\mathcal{I}}\}$, $|Q^{\text{fixed}}| = 10$.
\For{$j=1, \dots, M$} \Comment{$M$ trial attempts}
    \For{$i \in I$}
        \State Retrieve a sample $(q_{i}, a^{\text{CoT}}_{i}) \in \mathcal{X}$
        \State Sample $Q^{\text{random}}\gets \{(q_{k}, a_k^{\text{CoT}},  a^{\text{Prolog}}_{k})_{k\notin\mathcal{I}}\}$, $|Q^{\text{random}}| = 10$ from $\mathcal D$
        \State{// Adding 10 dynamic samples and 10 fixed samples into the 20-shot prompt.}
        \State Prompt GPT-4 with $\{q^{gen}\} \cup Q^{fixed} \cup Q^{random} \cup \{ q_{i}, a^{\text{CoT}}_{i}\}$ to obtain $a_i^{\text{Prolog}}$ 
        \If{$\mathcal{P}(a_i^{\text{Prolog}}) = \mathcal{C}(a^{\text{CoT}}_{i})$}
        \State $\mathcal{D} \gets \mathcal{D} \cup \{(p_{\text{ins}}, q_{i}, a_i^{\text{Prolog}})\}$
        \State $\mathcal{I} \gets \mathcal{I} \setminus \{i\}$
        \EndIf
    \EndFor
\EndFor
\If{$\mathcal{I} \neq \emptyset$}
    \State Manually correct Prolog codes $\{a_i^{\text{Prolog}}\}_{i\in\mathcal I}$ that solve $\{q_i\}_{i\in\mathcal{I}}$ 
    \State $\mathcal{D} \gets \mathcal{D} \cup \{(p_{\text{ins}}, q_{i}, a_i^{\text{Prolog}})_{i\in\mathcal{I}}\}$
\EndIf
\end{algorithmic}
\end{algorithm}
\subsection{Permutation procedures} \label{apdx:permute}
Permutations can be performed both on the level of facts or rules and on the level of goals in a rule. In practice, for each piece of code, we first permute the goals in the \texttt{solve<answer>:-<goal\_1>,<goal\_2>, ...} predicate. Since the total number of permutations is sensitive to the number of goals and can easily grow to a large magnitude, thus running out of memory, we used the \text{permutation} method in the \text{itertools} library to yield an iterator over the permutations. Then, we took up to 10 goal permutations from the iterator. If there were less than 10 goal permutations in total because the code was concise and there were not many goals, we took as many goal permutations as possible. Then, in the same manner, we took up to 10 fact and rule permutations. In principle, there would be at most 100 permuted samples generated for one original sample. Then, for each sample, while conducting an experiment that required a certain number of permutations, we randomly sampled permutations from the set of permutations of size up to $100$. For some sample, if the target number of permutations exceeded the total permutations it had, we took all its permutations instead.

\subsection{Instruction Prompt Samples} \label{sec:ins_prompt}
Below are the instruction prompts we used for different training settings (CoT, Prolog, and Permuted Prolog).

\begin{table}[H]
    \centering
    \footnotesize
    \begin{tabular}{p{4cm}|p{11cm}}
    \toprule
    Setting & Prompt Template\\ \midrule
    CoT & Below is an instruction that describes a task, paired with an input that provides further context. Write a response that appropriately completes the request.\\
        & \#\#\# Instruction:\\
        & Please generate an explanatory answer to solve the given math problem.\\
        & \#\#\# Input:\\
        & <Question>\\
        & \#\#\# Output:\\
        & <CoT Reasoning>\\\midrule
    Prolog Generation & Below is an instruction that describes a task, paired with an input that provides further context. Write a response that appropriately completes the request.\\
                      & \#\#\# Instruction:\\
                      & Please generate a piece of Prolog code to solve the given math problem.\\
                      & \#\#\# Input:\\
                      & <Question>\\
                      & \#\#\# Output:\\
                      & <Prolog Code>\\ \midrule
    Permuted Prolog Generation & Below is an instruction that describes a task, paired with an input that provides further context. Write a response that appropriately completes the request.\\
                               & \#\#\# Instruction:\\
                               & Please generate a piece of Prolog code in non-sequential order to solve the given math problem.\\
                               & \#\#\# Input:\\
                               & <Question>\\
                               & \#\#\# Output:\\
                               & <Prolog Code>\\
    \bottomrule
    \end{tabular}
\end{table}

\subsection{Error Analysis} \label{sec:err}
In this section, we present some typical errors made by our best-performing model to understand the behavior and bottleneck of LLM generation of prolog programs to solve arithmetic reasoning tasks. We sampled 70 problematic Prolog codes generated by finetuned Mistral using the 1:1 permutation setting, which had yielded the highest accuracy, that had errors and manually checked the type of errors using the Swi-Prolog interpreter. $100\%$ of them had semantic errors, meaning their logic of solution was wrong. $7\%$ of them had syntax errors, meaning the Swi-Prolog yields syntax errors on those codes. Some examples of typical syntax errors are listed below. The lines causing errors are in bold. The details of the errors are explained in the comments.

\begin{table}[H]
    \centering
    \scriptsize
    \begin{tabular}[t]{p{4cm}|p{11cm}}
    \toprule
    \textbf{Question} & \textbf{Prolog Code}\\ \midrule

    Julia's boat sprang a leak while she was out on the lake. The boat was taking on two liters of water for every ten feet she rowed back towards shore. It took her sixteen seconds to row twenty feet. The shore was 64 seconds away. How much water had the boat taken on by the time Julia reached shore?  & \begin{tabular}[c]{@{}l@{}}:- use\_module(library(clpq)).\\ \\ liters\_per\_10\_feet(julia, 2).\\ \\ time\_per\_20\_feet(julia, 16).\\ \\ distance\_to\_shore(julia, 64).\\ \\ solve(Water\_taken\_on) :-\\     liters\_per\_10\_feet(julia, Liters\_per\_10\_feet),\\     time\_per\_20\_feet(julia, Time\_per\_20\_feet),\\     distance\_to\_shore(julia, Distance\_to\_shore),\\     \{Seconds\_per\_10\_feet = Time\_per\_20\_feet / 2\},\\     \{Total\_seconds = Distance\_to\_shore / 10 * Seconds\_per\_10\_feet\},\\     \{Total\_feet\_rowed = Total\_seconds / \textbf{Seconds\_per\_20\_feet} * 20\},\ \textit{\% The variable has never been defined.}  \\     \{Total\_liters\_taken\_on = Total\_feet\_rowed * Liters\_per\_10\_feet\},\\     \{Water\_taken\_on = Total\_liters\_taken\_on\}.\end{tabular}\\

    \midrule

    Bill bakes 300 rolls, 120 chocolate croissants, and 60 baguettes every day. Each roll is 4 inches long, each croissant is 6 inches long, and each baguette is two feet long. If Bill puts all the baked goods end to end, how long will they be in feet? & \begin{tabular}[c]{@{}l@{}}:- use\_module(library(clpq)).\\ \\ bake(bill, rolls, 300).\\ bake(bill, chocolate\_croissants, 120).\\ bake(bill, baguettes, 60).\\ \\ \textbf{length}(roll, 4).\ \textit{\# The predicate has already been reserved by Prolog.}\\ \textbf{length}(chocolate\_croissant, 6).\\ \textbf{length}(baguette, 2).\\ \\ solve(Total\_length) :-\\     bake(bill, rolls, Num\_rolls),\\     bake(bill, chocolate\_croissants, Num\_chocolate\_croissants),\\     bake(bill, baguettes, Num\_baguettes),\\     length(roll, Roll\_length),\\     length(chocolate\_croissant, Chocolate\_croissant\_length),\\     length(baguette, Baguette\_length),\\     \{Roll\_total\_length = Num\_rolls * Roll\_length\},\\     \{Chocolate\_croissant\_total\_length = Num\_chocolate\_croissants * Chocolate\_croissant\_length\},\\     \{Baguette\_total\_length = Num\_baguettes * Baguette\_length\},\\     \{Total\_length = Roll\_total\_length + Chocolate\_croissant\_total\_length + Baguette\_total\_length\}.\\ \textit{\% This code also contains a semantic error because it fails to convert the unit of inch to feet.}\end{tabular}\\

    \midrule

    In a race with 240 Asians, 80 were Japanese, and the rest were Chinese. If the number of boys on the Chinese team was 60, how many girls were on the Chinese team? & \begin{tabular}[c]{@{}l@{}}:- use\_module(library(clpq)).\\ \\ total\_participants(240).\\ japanese\_participants(80).\\ chinese\_boys(60).\\ \\ solve(Girls\_on\_chinese\_team) :-\\     total\_participants(Total\_participants),\\     \textbf{Japanese\_participants} ,\ \textit{\% The correct syntax:\ japanese\_participants(Japanese\_participants),}\\     \{Chinese\_participants = Total\_participants - Japanese\_participants\},\\     chinese\_boys(Chinese\_boys),\\     \{Girls\_on\_chinese\_team = Chinese\_participants - Chinese\_boys\}.\end{tabular}\\

    \midrule
    
    Brinley is in Mr. Bert's math class. Mr. Bert gives six tests each semester. He allows the students to remove the lowest score from the average each semester. If Brinley has scores of 89, 71, 92, 100 and 86 on the first five tests, what score does she need on the sixth test to get an average of 93? & \begin{tabular}[c]{@{}l@{}}:- use\_module(library(clpq)).\\ \\ num\_tests(mr\_bert, 6).\\ lowest\_score\_removal(mr\_bert, 1).\\ target\_average(mr\_bert, 93).\\ \\ scores(brinley, {[}89, 71, 92, 100, 86{]}).\\ \\ solve(Test\_score) :-\\     num\_tests(mr\_bert, Num\_tests),\\     lowest\_score\_removal(mr\_bert, Lowest\_score\_removal),\\     target\_average(mr\_bert, Target\_average),\\     scores(brinley, Scores),\\     Length is Num\_tests - Lowest\_score\_removal,\\     \{Total\_score = \textbf{sum}(Scores)\},\ \textit{\% The built-in predicate is misused.}\\     \{Average\_score = Total\_score / Length\},\\     \{Test\_score = (Target\_average * Length) - Total\_score\}.\end{tabular}\\
    
    \bottomrule
    \end{tabular}
\end{table}

\subsection{Training Details and Computational Budget} \label{sec:gpu}
During finetuning, we controlled the number of epochs to be 6, batch size to be 128, and learning rate to be $3\times10^{-4}$. For a single training run, we used 2 NVIDIA RTX 4090 GPUs to finetune Llama2 and CodeLlama and 2 NVIDIA RTX 8000 GPUs to finetune Mistral. We adopted Distributed Data Parallelism to speed up training. Training on the original CoT data in GSM8K or the non-permuted Prolog code data took around 2 hours on 2 NVIDIA RTX 4090 GPUs and around 10 hours on 2 NVIDIA RTX 8000 GPUs. When we added in permuted samples, the training time grew proportionally with the dataset size since we controlled the number of epochs and batch size. During inference on the test set, we used a batch size of 2 on an RTX 4090 GPU, which took around 6 hours to finish a full inference round, and a batch size of 3 on one RTX 8000 GPU, which took around 7 hours to finish a full inference round.

\end{document}